\documentclass{article}

\usepackage{blindtext} % Package to generate dummy text throughout this template 
\usepackage{graphicx}
\usepackage{float}
\usepackage{amssymb}
\usepackage[sc]{mathpazo} % Use the Palatino font
\usepackage[T1]{fontenc} % Use 8-bit encoding that has 256 glyphs
\usepackage{microtype} % Slightly tweak font spacing for aesthetics

\usepackage[english]{babel} % Language hyphenation and typographical rules

\usepackage[hmarginratio=1:1,top=32mm,columnsep=20pt]{geometry} % Document margins
\usepackage[hang, small,labelfont=bf,up,textfont=it,up]{caption} % Custom captions under/above floats in tables or figures
\usepackage{booktabs} % Horizontal rules in tables

\usepackage{lettrine} % The lettrine is the first enlarged letter at the beginning of the text

\usepackage{enumitem} % Customized lists
\setlist[itemize]{noitemsep} % Make itemize lists more compact

\usepackage{abstract} % Allows abstract customization
\usepackage{titlesec} % Allows customization of titles
\titleformat{\section}[block]{\large\scshape\centering}{\thesection.}{1em}{} % Change the look of the section titles
\titleformat{\subsection}[block]{\large}{\thesubsection.}{1em}{} % Change the look of the section titles

\usepackage{fancyhdr} % Headers and footers
\usepackage{titling} % Customizing the title section

\usepackage{hyperref} % For hyperlinks in the PDF

\pretitle{\begin{center}\Huge\bfseries} % Article title formatting
\posttitle{\end{center}} % Article title closing formatting
\title{Generating Multilingual Parallel Corpus Using Subtitles} % Article title
\author{%
\textsc{\href {http://farshadjafari.ir}{Farshad Jafari}} \\[1ex] % Your name
\normalsize Department of Computer Science, Amirkabir University of Technology, Tehran, Iran \\ % Your institution
\normalsize \href{mailto:frshd.jfri@gmail.com}{frshd.jfri@gmail.com} % Your email address
}
\date{\today} % Leave empty to omit a date
\renewcommand{\maketitlehookd}{%
\begin{abstract}
\noindent Neural Machine Translation with its significant results, still has a great problem: lack or absence of parallel corpus for many languages. This article suggests a method for generating considerable amount of parallel corpus for any language pairs, extracted from open source materials existing on the Internet. Parallel corpus contents will be derived from video subtitles. It needs a set of video titles, with some attributes like release date, rating, duration and etc. Process of finding and downloading subtitle pairs for desired language pairs is automated by using a crawler. Finally  sentence pairs will be extracted from synchronous dialogues in subtitles. The main problem of this method is unsynchronized subtitle pairs. Therefore subtitles will be verified before downloading. If two subtitle were not synchronized, then another subtitle of that video will be processed till it finds the matching subtitle. Using this approach gives ability to make context based parallel corpus through filtering videos by genre. Context based corpus can be used in complex translators which decode sentences by different networks after determining content’s subject. Languages have many differences in their formal and informal styles, including words and syntax. Other advantage of this method is to make corpus of informal style of languages. Because most of movies dialogues are parts of a conversation. So they had informal style. This feature of generated corpus can be used in real-time translators to have more accurate conversation translations.

\end{abstract}
Keywords: Neural machine translation , Low-resource languages, Parallel Corpus generator
}

\begin{document}

% Print the title
\maketitle

%----------------------------------------------------------------------------------------
%	ARTICLE CONTENTS
%----------------------------------------------------------------------------------------

\section{Introduction}

Natural Language Processing is one of the most useful fields of computer science and machine translation is one of its branches. Machine translation is aimed at finding the right meaning of a sentence in source language, and express that meaning with the right syntax in destination language. This problem seems to be an NP problem in computer science. Lot of approaches has been designed to solve this problem which are classified in three categories: classical methods, statistical methods and neural methods. This article will discuss about approaches to improve results of statistical and mainly neural methods. Neural machine translation is a cutting edge technique which takes advantage of deep learning. By suggesting intelligent methods like sequence to sequence learning \cite{sutkever}, machine translation made lots of improvements in results.
Like all other deep learning systems, neural machine translation needs a huge amount of verified resources to train a neural network. These resources called parallel corpus. They contain number of translated sentence pairs in two languages. Parallel corpus is not present for many languages. Other languages have made their corpus from news contents, parliament texts or telegraph messages which their translation process were done for other purposes than machine translating. So implementing these procedures for a low resource language costs a lot of time and money. Computer scientists have suggested many approaches facing this problem. For example recent works about using similarities between a high-resource language and a low-resource language in order to transform accessible language into
low resource language has been done \cite{Karakanta2017}. This method achieved significant results without using a parallel corpus in low-resource language. This article discusses about a technique to collecting resources for all languages, which will produce more data over time.
In 2017, about 700 movies has been produced in US (s.follows, 2017 ). For almost all of the movies, there exist many subtitles for any language. These subtitles are nothing but parallel corpus in machine translation context, and this article will describe a method for generating parallel corpus using video subtitles. Using these resources will give us some facilities like providing data for low-resource languages, informal style of language translation and context-based translation. 
This article contains 5 parts. Part 2 discusses about algorithms and procedures of this method. Part 3 provides a documentation for implemented framework using this method. Part 4 describes the results of framework as a case study, and finally part 5 have some conclusions about the method applications and possible improvements.

%------------------------------------------------

\section{Method}

This method contains following parts:
\begin{itemize}
\item providing a list of videos
\item crawling video subtitles for a language pair
\item extracting synchronous dialogues from subtitle pairs
\item deriving sentence pairs from dialogue pairs
\item generating parallel corpus using sentence pairs
\end{itemize}
which are discussed in detail in following sections.

%------------------------------------------------
\subsection{List of videos}

First of all we need a list of videos that may have texts, like movies, series and etc. Without such dataset we are unable to crawl and classify contents we need. 

\subsubsection{source}
There are many ways to achieve our first goal. One is writing a crawler to extract video informations from online movie databases like imdb. But with a little bit of searching on the Internet, we can find such generated datasets. One we use in the framework is uploaded at kaggle, that has more than 14000 indexes crawled from imdb. 

\subsubsection{filters}
The required dataset is better to have some information about each movie. Because later on by using limits on each attribute we can make more specific corpus from their subtitles. 
Our applied dataset includes following attributes:
title, words in title, imdb rating, rating count, duration, year, type, and the rest of the fields are 0/1 variables indicating if the movie has the given genre like Action, Adult, Adventure and so on.

\subsubsection{storage}
Using a database engine for processing each part of solution will help a lot. Some useful tasks like logging, debugging and further processes can be made easily by a database back-end. So the framework uses mongodb(a no-SQL database system). 
For starters, we store our movie dataset in database.

\subsection{Subtitle Crawler}
In order to download subtitles we need a crawler which takes video title and language as input and downloads the proper subtitles from open source subtitle databases. 
Its hard to write such crawler in short time. Again with a bit of searching the Internet we can find such packages which will meet our needs. One we used in our framework is a nice python package named subliminal that has a full api documentation. 

\subsubsection{synchronized subtitles}
Next step is facing time alignment problem in subtitles, because there are a lot of editions for each video with different timings. 
There exist many ways to handle it. One is matching dialogue pairs with word to word translation. It has a lot of disadvantages. First, subtitles are using informal language which is hard to translate with existing dictionaries. Second, this approach will bind our framework to a lot of dictionaries for each language pair.
After a hard try implementing this, final matcher reached to average of 10 percent occurrence at each matching dialogue.
The other solution is to find subtitles with exact timing. Because subtitle translators usually prefer to edit an existing subtitle file in source language to generate new subtitle in destination language. This solution will ensure each subtitle pair to be synchronized before downloading. For unsynchronized subtitles it will process another source or destination subtitle till it finds the right one. That is easy to implement and reliable. So our framework is using it. 
The better but harder solution is shifting subtitles time-line to match each other. This can be done in next versions of framework.

\subsection{Sentence Pairs}
Easy but risky task is to extract sentence pairs from dialogue pairs.  
Because there are lot of possibilities in a dramatic dialogue. For example interjections, half sentences, multi-sentence dialogues and so on.

\subsubsection{cleaning dialogues}
Before storing synchronous dialogues in database, they need to be cleaned from extra punctuations, color tags, descriptions, etc. 

\subsubsection{matching sentences}
Sentence pairs are extracted from synchronous dialogue pairs. There are three possibilities for this step. 
\begin{itemize}
\item Each dialogue contains one or half sentence. Which is the best event, we store them as they are.
\item Each dialogue pair has multiple sentences, with equal number in source and language. This event is also harmless, so we split dialogues in sentences and store them with same order.
\item Each dialogue pair has multiple sentences, with different number in source and language. In this rare situation best approach is to skip dialogues or just store equal number of sentences from each dialogue.
\end{itemize}

\subsection{Corpus files}
The last and and obvious step is to generate corpus files from sentence pairs stored in db.

\subsubsection{related application formats}
Machine translation frameworks have their own formats of input training file. For example tensor-flow, keras and theano have same parallel corpus formats. They need two files containing sentences in each line, matching their translation in other file at the same line.

\subsubsection{cleaning}
Another step of cleaning is needed to handle mistakes occurred in previous steps like empty strings, punctuation-only lines, etc. These errors are inevitable. Because these contents are generated by a large number of translators with different styles and habits.

\section{Manual}
Using the described method we implemented a small version of working framework to examine its results. The code is open-source and available to use. It named Parallel Corpus Generator, written in python language, using mongodb as database. Therefore  working with it needs a fair knowledge of python language. 
For running project do the following steps: 
\begin{itemize}

\item Install python on your system if dont have it. More information about installing python can be found \href{https://www.python.org/downloads/}{here}.
\item Install and run mongodb. More information about installing mongodb can be found \href{https://docs.mongodb.com/manual/installation/}{here}.
\item clone the project from \href{https://github.com/farshadjafari/parallel_corpus_generator.git}{github repository}.
\item create a virtual environment for python somewhere near the code:
\begin{verbatim}
virtualenv venv
source venv/bin/activate
\end{verbatim}
\item install required python packages with pip:
\begin{verbatim}
pip install requirements.txt
\end{verbatim}
\end{itemize}
you can run all steps with default configuration by running following command, but it may take hours to days to get done. I recommend to run each step separately:
\begin{verbatim}
python main.py {source language code} {destination language code}
\end{verbatim}
\footnote{replace  source and destination language code with your desired alpha2 language codes. For example:
python main.py en fa}

\subsection{Initializing videos}
in order to storing movie dataset in database run following command:
\begin{verbatim}
python movies/reader.py {source language code} {destination language code}
\end{verbatim}

\subsection{Filters}
edit movies/filter.py file for desired movie filters. 
The number of videos after filtering will show by running:
\begin{verbatim}
python movies/filter.py
\end{verbatim}

\subsection{Crawl subtitles}
in order to crawl and download synchronized subtitles for desired filters and languages run:
\begin{verbatim}
python downloader/main.py {source language code} {destination language code}
\end{verbatim}

\subsection{Store Dialogues}
in order to storing synchronous dialogues in database run:
\begin{verbatim}
python subtitle/main.py {source language code} {destination language code}
\end{verbatim}

\subsection{Match sentence pairs}
in order to extract sentence pairs from stored dialogues run:
\begin{verbatim}
python matcher/main.py {source language code} {destination language code}
\end{verbatim}

\subsection{Generate corpus}
finally after running all commands above for generating parallel corpus run:

\begin{verbatim}
python generator/main.py {source language code} {destination language code}
\end{verbatim}

\section{Case Study}
We have trained a tensorflow-nmt with english-persian parallel corpus generated by this method. In a few amounts of sentences and learning steps we gained an acceptable (but not applicable )results which is up here. This section describes its detail and statistics.

\subsection{Filters count}
\begin{verbatim}
TYPE_LIMIT = 'video.movie'
YEAR_LIMIT = 1960
RATE_LIMIT = 6
DURATION_LIMIT = 3600
\end{verbatim}

\begin{table}[H]
\caption{Number of Filtered Movies}
\centering
\begin{tabular}{lr}

Filters & Movies count \\
\midrule
all videos & 12083 \\
all movies & 9865 \\
year >= 70`s movies & 8286 \\
after 70`s movies with Imdb rating >= 6 & 6531 \\
after 70`s movies min 6 rating duration >= 60 & 6158 \\ 

\bottomrule
\end{tabular}
\end{table}

\subsection{Found subtitle pairs}
\begin{table}[H]
\caption{Found Subtitle Pairs Stat}
\centering
\begin{tabular}{lr}
\midrule
subtitle pairs count & 541 \\
Time spent & $\simeq$ 20 hours \\
avg time finding each pair & 541/1200 $\simeq$ 2 minutes \\
\bottomrule
\end{tabular}
\end{table}

\subsection{Found dialogues}
\begin{table}[H]
\caption{Found Dialogues Stat}
\centering
\begin{tabular}{lr}
\midrule
Synchronous Dialogues count & 644284 \\
avg dialogues count per movie & 644284/541$\simeq$1190 \\
\bottomrule
\end{tabular}
\end{table}

\subsection{Found sentences}
\begin{table}[H]
\caption{Found Sentences Stat}
\centering
\begin{tabular}{lr}
\midrule
sentence pairs count & 682129 \\ 
Avg sentence pair  per movie & 682129/541$\simeq$1260 \\
Number of sentences per dialogue & 682129/644284$\simeq$1.05 \\

\bottomrule
\end{tabular}
\end{table}

\subsection{Found sentences}
We trained tensorflows nmt with $\href{https://github.com/tensorflow/nmt/blob/master/nmt/standard_hparams/wmt16_gnmt_4_layer.json}{wmt16 gnmt 4 layer}$ as hparams, except with 512 number of units. Our machine's specs was:

\begin{table}[H]
\caption{Computer's spec}
\centering
\begin{tabular}{lr}
\midrule
CPU Core i7 \\
RAM & 16 GB DDR4 \\
GPU memory & 4 GB \\
GPU Compute Capability & 5.0 \\
Time Spend for 60000 training steps & $\simeq$ 15 hour \\
\bottomrule
\end{tabular}
\end{table}

\subsubsection{Training results}

\begin{table}[H]
\caption{Training Stat}
\centering
\begin{tabular}{lr}
\midrule
dev bleu (60k steps) & 6.862 \\
dev perplexity (60k steps) & 35.56 \\
train perplexity (60k steps) & 12.81 \\
train loss (60k steps) & 29.26 \\
best bleu (53k steps) & 7.073 \\
\bottomrule
\end{tabular}
\end{table}

\begin{figure}[H]
\centering
\includegraphics[width=\textwidth]{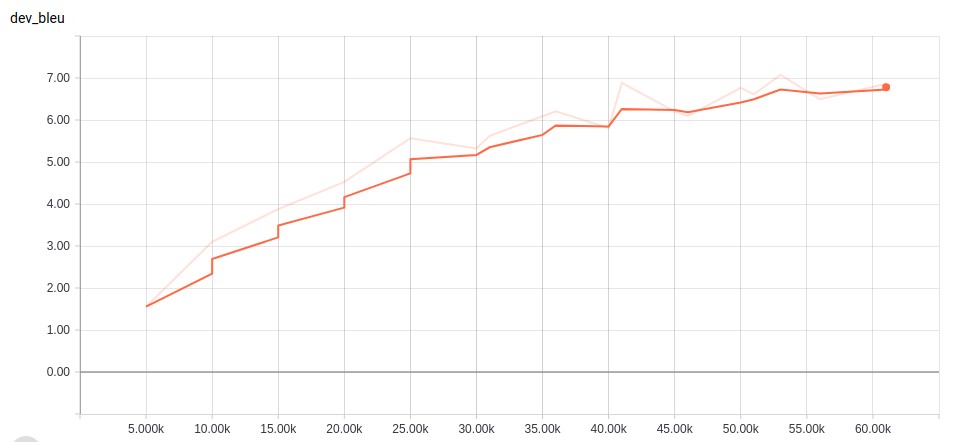}
\caption{Dev Bleu}\label{visina8}
\end{figure}

\begin{figure}[H]
\centering
\includegraphics[width=\textwidth]{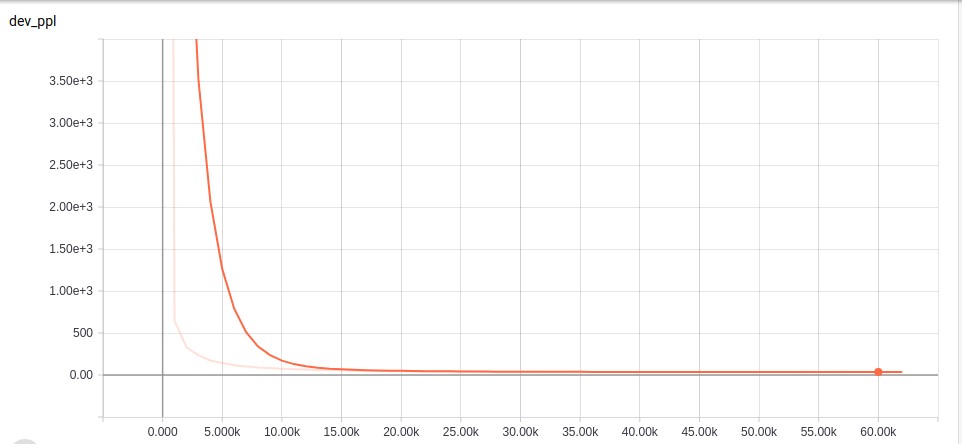}
\caption{Dev ppl}\label{visina8}
\end{figure}

\begin{figure}[H]
\centering
\includegraphics[width=\textwidth]{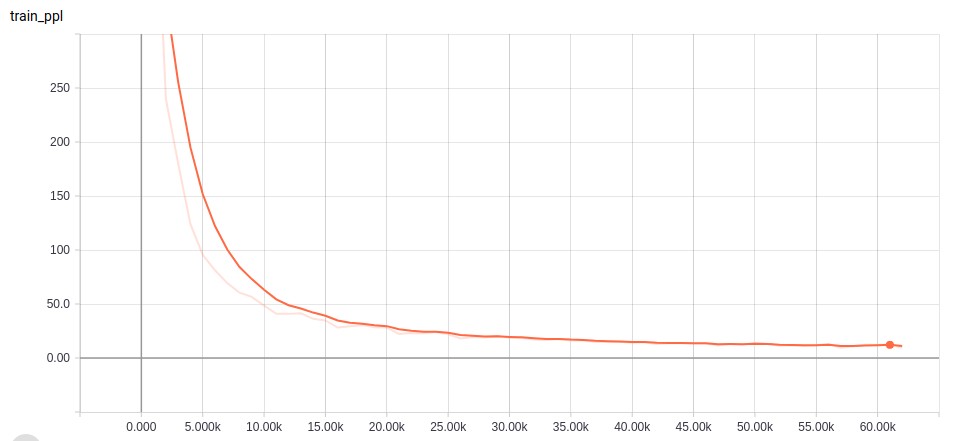}
\caption{Train ppl}\label{visina8}
\end{figure}

\begin{figure}[H]
\centering
\includegraphics[width=\textwidth]{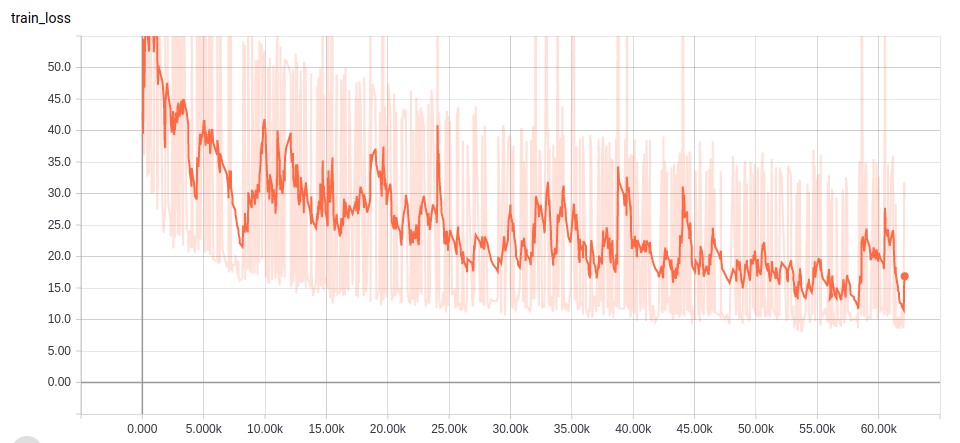}
\caption{Train Loss}\label{visina8}
\end{figure}

\section{Conclusion}
Like any other solutions for this problem, this method has its cons and pros. It may have a lot of deficiencies which is not covered in this article. But I believe that it worth to work on it with more effort. It has a number of applications which are not implemented yet, and a lot of improvements that can increase the quality of neural machine translators. In the following parts we will discuss some of them.

\subsection{Applications}
\subsubsection{Informal and spoken language translation}
As we know, most of the movies and series are written and acted by informal language. Existing translators are trained using formal language style of newspapers, parliament talks or wikipedia. But the main and next usage of translators is to provide better translations for spoken language. Because transportation is getting easier, voice recognition systems having impressive improvements and social networks are growing continuously. Therefore machine translation must make international communications easier.

\subsubsection{Intelligent assistants}
Intelligent assistants are designed to communicate with humans by natural language. The most annoying issue of trending assistants is their disability to have a normal conversation with human being. By using corpus generated with this method we can train them to make a dialogue. We can simulate the logic of conversation by training them on sequence of sentences. Chat bots are best describing examples of this subject. Even in future they may pass the Turing test.

\subsubsection{Context based translation}
If we learn separate networks using different genres of movies, we will have a network of translators. By adding a classifier besides them which determines the context of source language, then we will have more accurate translation in destination language using the same context translator.

\subsubsection{Low resource languages}
Languages are organism. They may extinct without native speakers. Nowadays if a society can not communicate with others, slowly will replace parts of its language with a foreign vocab or grammar. So their language will die. Low resource languages need to be translated and must have more resources, in order to have more speakers. 
As I believe, the main international problem of these days is misunderstanding. So by providing such facilities for societies to communicate, we will hand over a better world to our next generation.

\subsection{Improvements}
\subsubsection{Context based corpus}
By importing genre data of videos in db and make proper query on a context we can exploit a more relevant corpus in a desired field. Which is not done yet.

\subsubsection{Subtitle time shifting}
On average, 10 percent of subtitles were fully aligned. By developing a simple shifting algorithm, we can expand our corpus size up to 10 times. Which is sufficient for learning a high level nmt system. 
This algorithm should take asynchronous subtitle pair, shift their time-line by adding milliseconds till it obtain a sufficient number of paired points.

%------------------------------------------------

%----------------------------------------------------------------------------------------
%	REFERENCE LIST
%----------------------------------------------------------------------------------------

%----------------------------------------------------------------------------------------

\end{document}